\documentclass[letterpaper, 10 pt, conference]{ieeeconf}  % Comment this line out if you need a4paper

\IEEEoverridecommandlockouts                              % This command is only needed if 
\usepackage{graphics}
\usepackage{graphicx}
\usepackage{pdfpages}
\usepackage{color}
\usepackage{amsmath,amssymb}
\usepackage{mathtools}
\usepackage{hyperref}
\usepackage{color, soul}
\usepackage{multirow}
\numberwithin{equation}{section}
\renewcommand\thesection{\arabic{section}}
\usepackage{grffile} %remove annoying figure name in the pdf.

\usepackage{algorithm,algcompatible,amsmath}
\algnewcommand\INPUT{\item[\textbf{Input:}]}%
\algnewcommand\OUTPUT{\item[\textbf{Output:}]}%
\algnewcommand\algorithmicreturn{\textbf{return}}
\algnewcommand\RETURN{\State \algorithmicreturn}
\usepackage{widetext}
\usepackage{comment}

\title{\LARGE \bf
Solving Vehicle Routing Problem for unmanned heterogeneous vehicle systems using Asynchronous Multi-Agent Architecture (A-teams)
}

\author{Subramanian Ramasamy$^{1\dagger}$, Md Safwan Mondal $^{2}$, Pranav A. Bhounsule $^{1}$% <-this % stops a space
\thanks{$^{1}$ Subramanian Ramasamy, $^{2}$ Md Safwan Mondal, and $^{1}$ Pranav A. Bhounsule are with the Department of Mechanical and Industrial Engineering, University of Illinois Chicago, IL, 60607 USA. 
  {\tt\small sramas21@uic.edu}   \hspace{2mm}
  {\tt\small pranav@uic.edu}  \hspace{2mm}   
  {\tt\small mmonda4@uic.edu}   
  $^\dagger$ This work was supported by ARO grant W911NF-14-S-003.
  }%
}

\begin{document}

\maketitle
\thispagestyle{plain}
\pagestyle{plain}

\begin{abstract}
Fast moving but power hungry unmanned aerial vehicles (UAVs) can recharge on slow-moving unmanned ground vehicles (UGVs) to survey large areas in an effective and efficient manner. In order to solve this computationally challenging problem in a reasonable time, we created a two-level optimization heuristics. At the outer level, the UGV route is parameterized by few free parameters and at the inner level, the UAV route is solved by formulating and solving a vehicle routing problem with capacity constraints, time windows, and dropped visits. The UGV free parameters need to be optimized judiciously in order to create high quality solutions. We explore two methods for tuning the free UGV parameters: (1) a genetic algorithm, and (2) 
Asynchronous Multi-agent architecture (A-teams). The A-teams uses multiple agents to create, improve, and destroy solutions. The parallel asynchronous architecture enables A-teams to quickly optimize the parameters.
Our results on test cases show that the A-teams produces similar solutions as genetic algorithm but with a speed up of 2-3 times.
\end{abstract}

\section{INTRODUCTION} \label{sec:Introduction}
There has been a considerable increase in the use of the small Unmanned Aerial Vehicles (UAVs) across different spectrums such as entertainment and logistics \cite{altshuler2018optimal}.  The reason for such a widespread adaption is because they are agile robots that can navigate at high speeds in complex environments otherwise inaccessible to humans \cite{song2022policy}. Although UAVs are fast, they are severely limited to relatively small area due their limited battery capacity \cite{theile2020uav}.

In order to realize the potential of UAVs for large scale targets coverage, they can be paired with Unmanned Ground Vehicles (UGV) to provide on demand recharging depots.  Such cooperative routing of a team of UAV-UGVs have been utilized in tasks such as monitoring and inspection in congested urban environments \cite{gao2020commanding}, general surveillance \cite{langerwisch2012realization}, and post-disaster relief \cite{lakas2018framework}. 

The cooperative routing of UAV-UGVs is complex and computationally challenging \cite{baldacci2008routing}. 
The formulation of the problem involves minimizing a cost such as the time or fuel consumption while constraining the fuel capacity and speed limits of the UAV and UGV and ensuring that they are able to rendezvous efficiently. Although it is relatively easy to formulate the problem, it is difficulty to solve the formulation using exact methods due to the combinatorial nature of the problem. However, using suitable heuristics, it is possible to achieve high quality solutions relatively quickly. 

%%%%%%%%%%%%%%%%%%%%%%%%%
\begin{figure*} [h]
\centering
\includegraphics[scale=0.475]{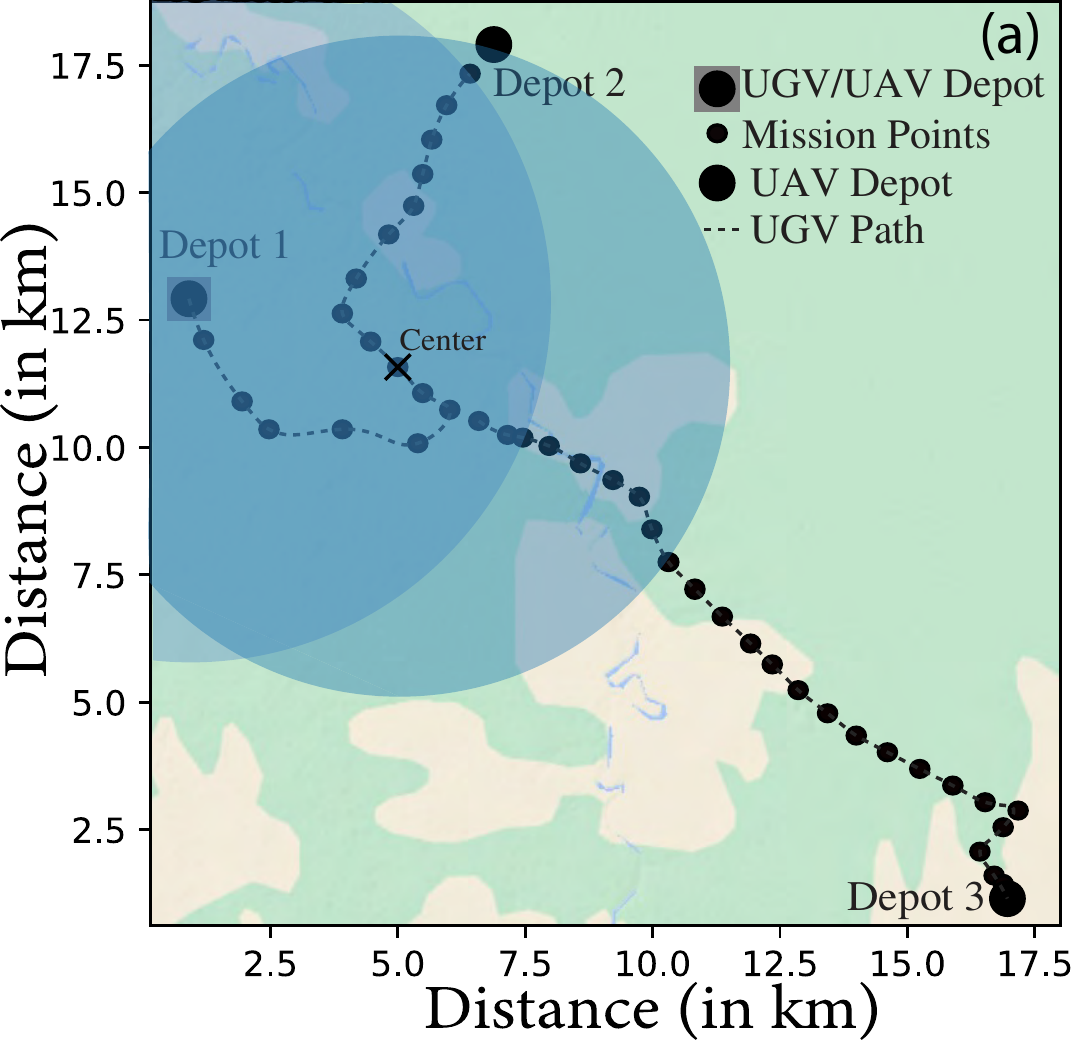}
\includegraphics[scale=0.475]{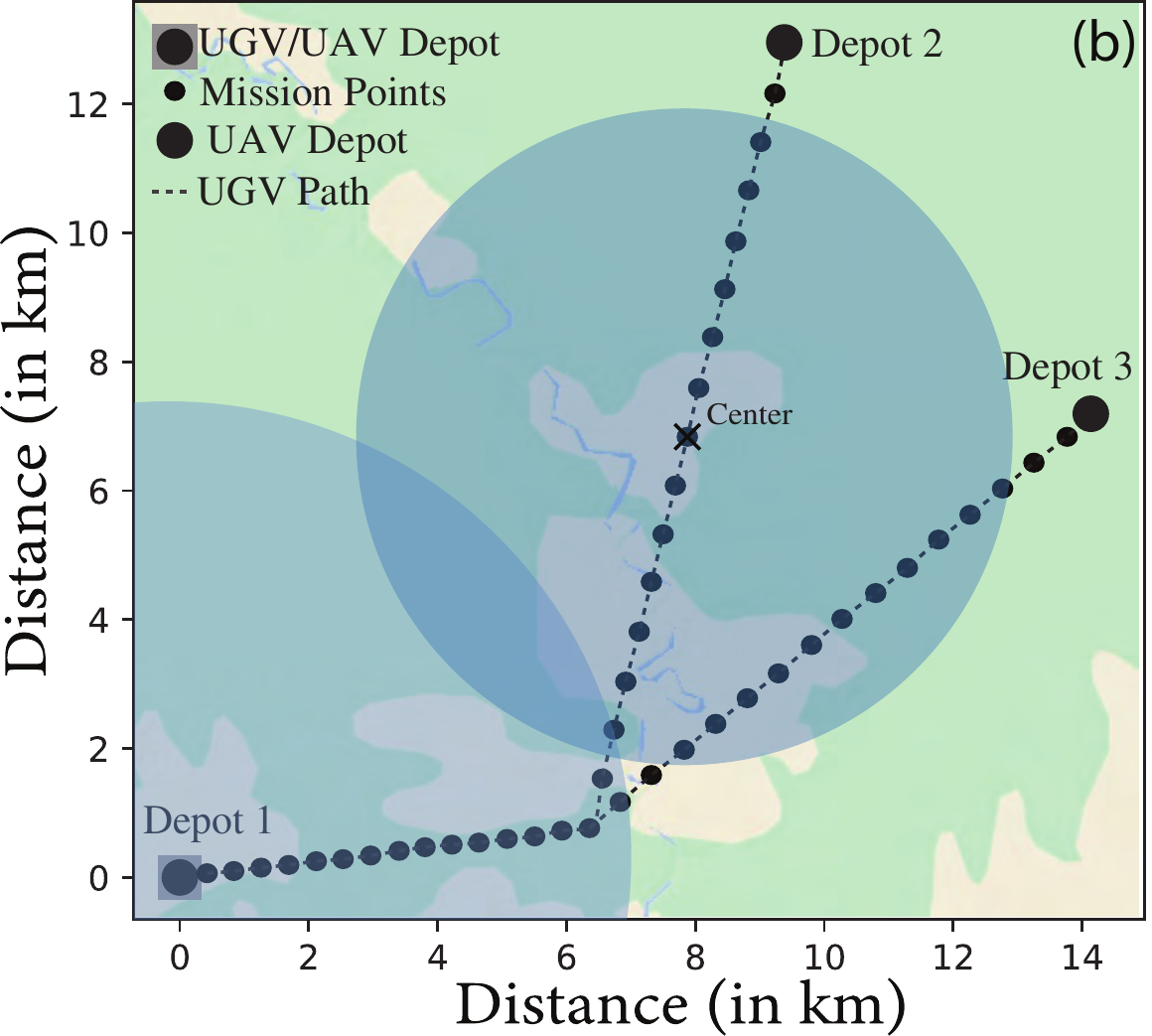}
\includegraphics[scale=0.475]{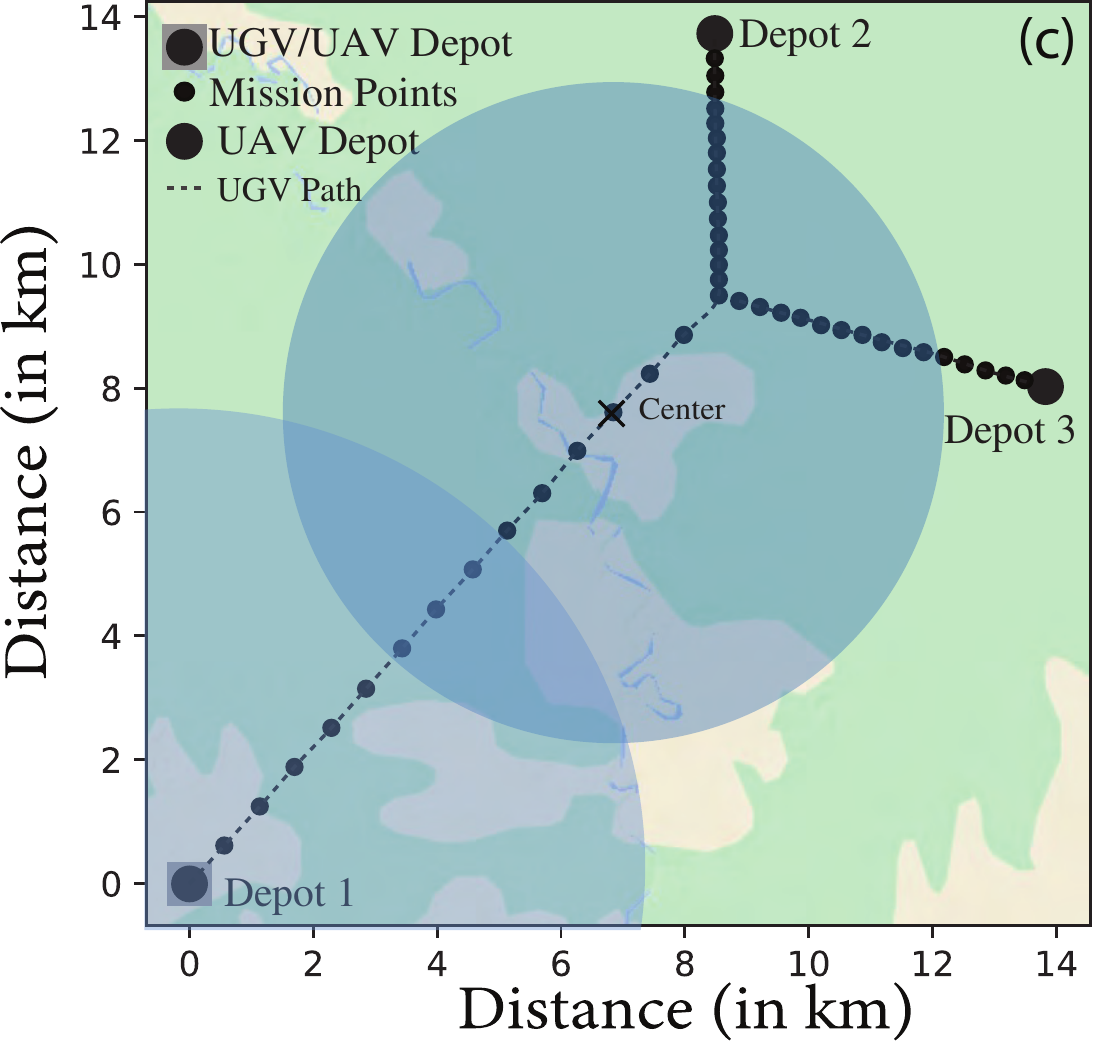}
\vspace{-0.25cm}
\caption{Description of different scenarios. The UAV and UGV, both start from one of the recharging depots. The target points are shown with black dots. The UAV range is shown with a blue circle. a) Scenario 1 b) Scenario 2 c) Scenario 3.}
\label{fig:scenario}
%\vspace{-0.75cm}
\end{figure*}
%%%%%%%%%%%%%%%%%%%%%%%

Figure~\ref{fig:scenario} shows the problem scenarios considered in this paper. The target points are shown with black dots. There are 3 recharging depots shown with a black dot that is bigger than the one used for target points. The UAV can travel on the UGV or fly by itself. The UAV may be charged by the UGV or at the depot. Both, UAV-UGV have to start and end at the depot. They both start and end at the same depot. %Furthermore, we assume that `Depot 1'  is always available to recharge both, the UAV and the UGV. However, `Depot 2' and `Depot 3' may be used for recharging the UAV only if the mission starts and ends at any one of them.

The blue circles represent the range of the UAV on a full charge; the distance that the UAV can cover is the diameter of the circle. For example, consider Figure ~\ref{fig:scenario} (a). If the UAV starts from the center of the circle on a full charge, it can return back to the center of the circle with an empty charge if it travels straight out and back. We have drawn two circles which are centered approximately at $(1,12.5)$ km and $(5,12)$ km. It can be observed that from the start location, the UAV cannot travel to the set of target points approximately from $(7.5,10)$ km to $(10,8)$ km. However, if the UAV starts from the point $(5,12)$ km with a full charge, it can cover those sets of target points and return back. To enable this solution, the UAV would need to ride with the UGV along the UGV till $(5,12)$ km, then visit the target points within that radius and get refueled. Meanwhile, the UGV stops at $(5,12)$ km location and waits for some time. Although, such a UGV stop helps to cover additional target points, there are some target points along the bottom right region that are left out. Hence, in such case, either the UGV has to have another stop along that region so that UAV can cover those target points and utilize that UGV stop to recharge or the UGV itself should travel along that path to cover all of those target points. From this figure, you can see that all 3 branches intersect at a common point $(6.1,10.8)$ km. Other scenarios are considered similar to the distribution of this scenario where three different branches meet at a point. This illustrates some of the intricacies of choosing an appropriate path for the UGV such that the UAV can successfully cover the target points at the extreme ends. This is an optimization problem where optimal routes are to be found for both UGV and UAV. In case of UGV, its route is modeled as a parameter set consisting of two UGV stop locations to recharge the UAVs, the wait time of UGV at those corresponding stops, and the starting or ending point of the entire route plan. The optimal solution corresponds to a UGV routes parameter set and its subjected UAV route for which the overall objective function is minimized.

There has been a considerable work done in the literature related to solving fuel-constrained routing of UAVs. Sundar et. al., \cite{sundar2013algorithms} worked on Fuel-Constrained UAV Routing Problem where a generalization of the asymmetric Traveling Salesman Problem (TSP) is solved using Approximation algorithm and fast heuristics. A Mixed Integer Programming Problem formulation is also proposed to obtain optimal solutions. Here, a single UAV is used and gets recharged on fixed depots. Venkatachalam et. al., \cite{venkatachalam2018two} modeled a multiple fuel-constrained UAV routing problem with fixed recharging depots. Here, the authors implemented a two-stage stochastic optimization problem with uncertainties in the fuel consumption of UAVs. Heuristics are used to achieve high quality solutions with faster computing time.

Some extensions in the aspect of heterogeneous vehicle routing were also considered in the literature to overcome some of the limitations existed in such fuel-constrained UAV routing problem with fixed depots. Subramanian et. al. \cite{ramasamy2021cooperative} considered the vehicle routing problem of multiple fuel-constrained UAVs and a single UGV that acts as a mobile recharging vehicle. The problem is being solved in a tiered fashion. The authors used K-means clustering and TSP to solve UGV routing problem, and then implemented Vehicle Routing Problem (VRP) with fuel, time and optional node constraints. The aforementioned work is extended in \cite{ramasamy2022coordinated} where a more generalized approach is taken to solve several different scenarios and proved the robustness of the algorithm. The above works allows the UGV to move freely on any paths, but there are some works in the literature that considers heterogeneous UAV-UGV vehicle routing with UGV constrained to move on certain prescribed paths. This is a challenging problem because each of these vehicles have different constraints on speed and fuel capacity. Maini et al. \cite{maini2018cooperative} considered the problem of routing a single fuel-constrained UAV to a set of target points while being recharged by stopping at a UGV traveling on a road network. They solved the problem using a two-stage approach. First, using the UAV range constraints, they found a set of recharging depots. Second, they formulated a mixed-integer linear program and solved for the path of both the UAV and UGV. The work in \cite{ramasamy2022heterogenous} also allows the UGV to move freely only on prescribed paths and is also followed in this paper.

Since UGV has a fixed route, the heuristics for the UGV could be modeled as a parameter tuning problem as that would provide a better solution by tuning the heuristic parameters. \cite{huang2019automatic} applied Bayesian Optimization to tune the hierarchical decomposition algorithm parameters and thus helped to achieve optimal solutions at a faster rate. Although such algorithms help us to achieve globally optimal solutions, they come at a cost of significant computational time. This was seen from the previous work by the authors \cite{ramasamy2022heterogenous} where Genetic Algorithm (GA) and Bayesian Optimization (BO) are implemented for parameter tuning to obtain the UGV route. The results from that work show that particularly in case of GA, higher computational time of about $180$ minutes was needed to solve that problem. The reason for high computational time is that these algorithms GA and BO are basically global optimization algorithms. Although such global optimization algorithms give search over a larger space, this compromises their efficiency. Hence they can be used when the computational time is not critical, such as offline optimization \cite{boyd2004convex}. At the same time, local optimization algorithms provide quicker solutions, but are optimal in a small region of space. 

To achieve faster global optimal solutions, Sachdev \cite{sachdev1998explorations} worked on proposing an architecture called A-Teams, which was originally developed by Talukdar et. al. \cite{talukdar1992cooperative}. In A-teams, global optimization methods search over a larger space to find potentially feasible solutions. These are then improved by the local optimization methods. The author implemented two algorithms, Stochastic Quadratic Programming (SQP), a local optimization method, and Genetic Algorithm (GA), a global optimization method, in A-Teams to show how the advantages of local and global optimization algorithms can be tapped to produce a better result in a computationally efficient manner. Jedrzejowicz et. al., \cite{jkedrzejowicz2014reinforcement} proposed A-Teams to solve a Resource Constrained Project Scheduling Problem. The authors perform Reinforcement Learning (RL) along with using other optimization algorithms like local search, tabu search to solve the problem using A-Teams. The RL component in their architecture helps to apply dynamic strategy for interaction between those different optimization algorithms in an A-Team. Those authors use a middleware called JABAT (Java Agent DEveleopment-Based A-Teams), to implement the A-Teams architecture. Kazemi et. al., \cite{kazemi2009multi} implemented A-Teams for solving a Production-Distribution Planning Problem where each agent in their architecture uses a GA sub-module to handle its tasks and conclude that the combined multi-agent GA system provides better solutions than the individual ones for their problem. Recent works by Jedrzejowicz et. al., \cite{jedrzejowicz2020experimental} involves implementing this architecture to solve a Resource Investment Problem in which different agents use Local search, Lagrangian relaxation, Path relinking algorithms, Crossover operators and cooperate together to solve such a problem.

The usage of A-Teams is also found amongst the routing problems in the literature. Rabak et. al. \cite{rabak2003using} presented the A-Teams framework to optimize the automatic electronic component insertion process on an inserting machine. They implemented a combination of Quadratic Assignment Problem and Traveling Salesman Problem (TSP) in the framework to perform optimization. The above work shows the framework's ability to handle multiple algorithms simultaneously. Such a realization becomes helpful in this work where the A-Teams come in hand for utilizing the local and global optimization algorithms, whose pros and cons were talked about a few lines before. Barbucha et. al., \cite{barbucha2013team} worked on investigating the effects and impact of a Team of A-Teams working in parallel to solve difficult combinatorial optimization problems. In their work, different algorithms cooperate together within an A-Team, and several similar A-Teams are made to work in parallel. The computational efficiency of their architecture is demonstrated by solving benchmark instances in different problems like Euclidean Planar Traveling Salesman Problem (TSP), Vehicle Routing Problem (VRP), Clustering Problem (CP), and Resource Constrained Project Scheduling Problem (RCPSP). Rachlin et. al., \cite{rachlin1999teams} implemented the A-Teams to solve a Traveling Salesman Problem (TSP) by using Farthest Insertion and Arbitrary Insertion heuristic algorithms in their architecture.

The A-teams has been limited to solving simple routing problems (e.g., TSP). Thus, the main contribution of this work is that we use the A-teams architecture to solve a heterogeneous and co-operative vehicle routing problem involving a UAV and a UGV. We also compare the A-teams architecture with results obtained using genetic algorithms in several scenarios. The flow of the paper is as follows. We present details about the optimization method in Sec.~\ref{sec:methods}. The results are in Sec.~\ref{sec:results}, followed by the Discussion in Sec.~\ref{sec:discussion}. Finally, the conclusion and future work is in Sec.~\ref{sec:conclude}

 %%%%%%%%%%%%%%%%%%%
\begin{figure}[t]
\centering
\includegraphics[scale=0.35]{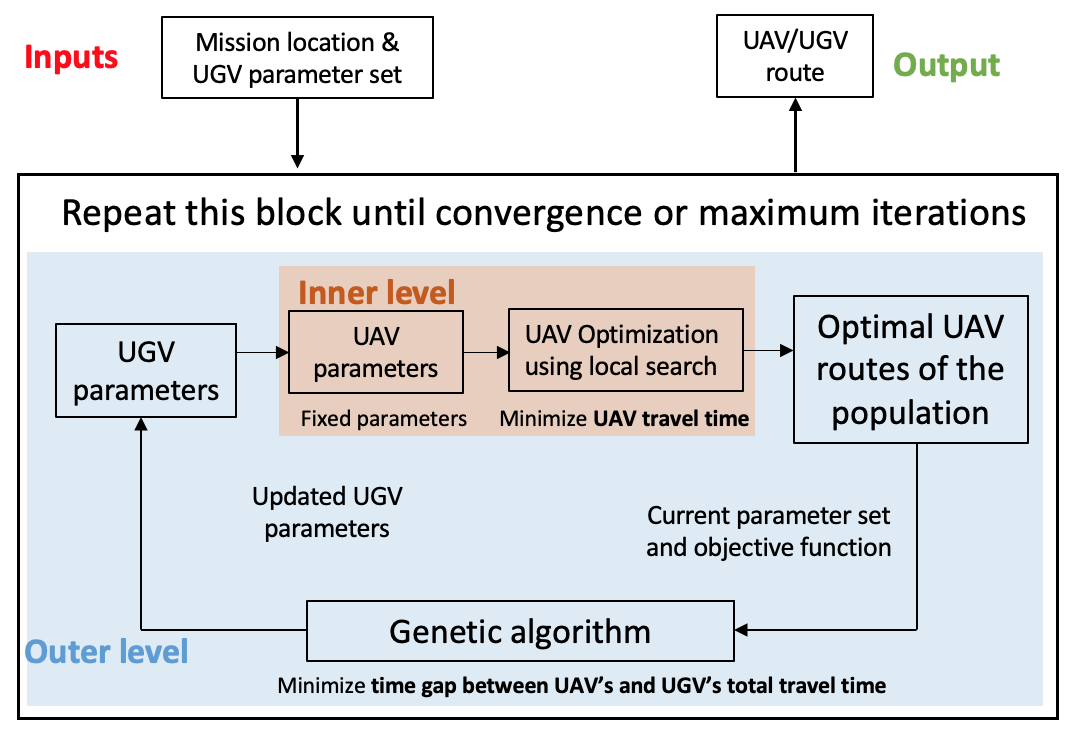}
\vspace{-0.25cm}
\caption{Overview of the algorithm. The outer-level block is run in parallel on multiple cores.}
\label{fig:algorithm}
\end{figure}
%%%%%%%%%%%%%%%%%%%

 %%%%%%%%%%%%%%%%%%%%%%%%%%%%%%%%
\begin{figure*}[t]
\centering
\includegraphics[scale=0.32]{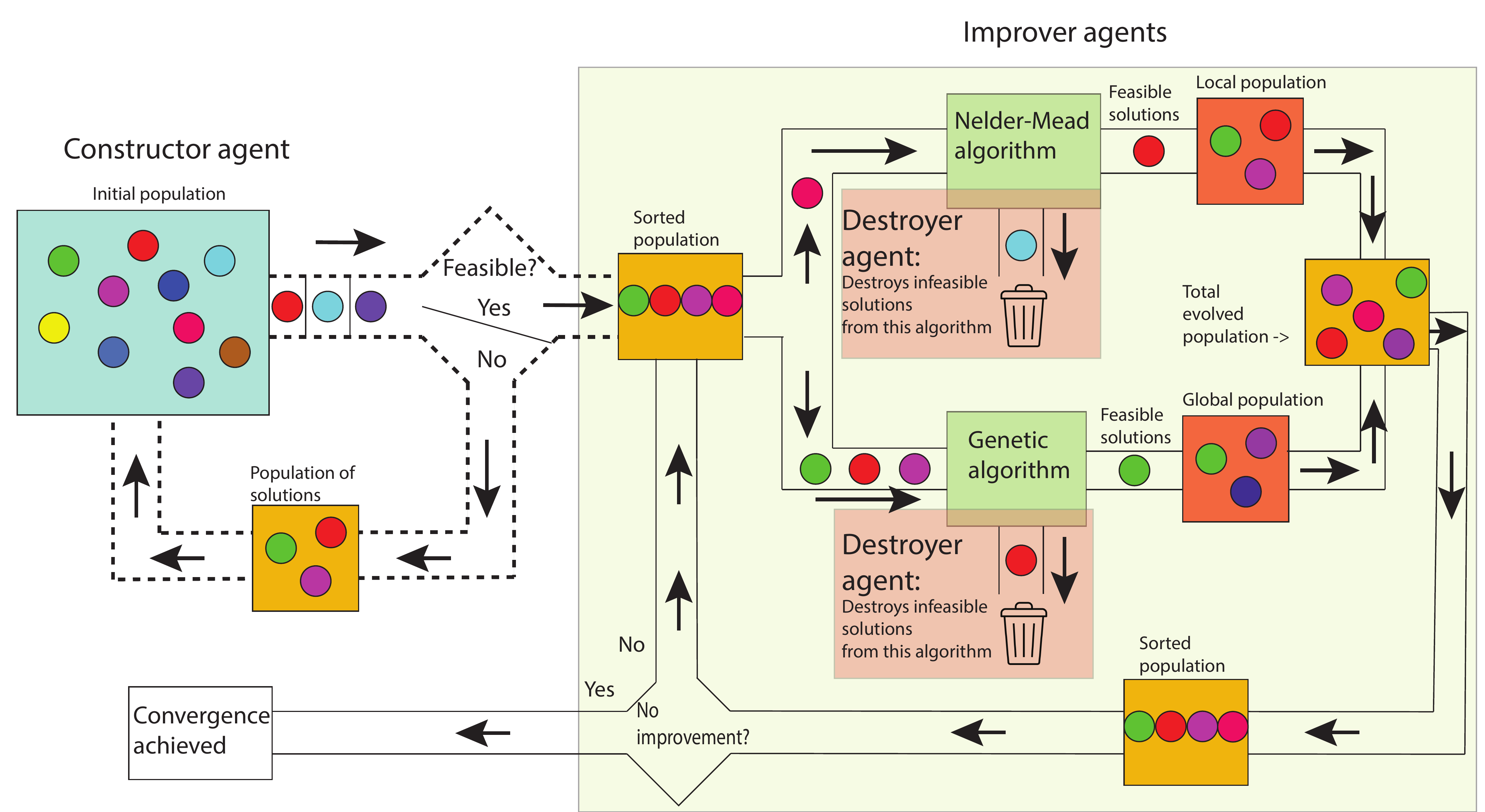}
\vspace{-0.2cm}
\caption{Implementation of A-Teams architecture for this cooperative routing problem.}
\label{fig:Architecture description}
%\vspace{-0.75cm}
\end{figure*}
%%%%%%%%%%%%%%%%%%%%%%%%%%%%%%%%

%%%%%%%%%%%%%%%%%%%%%%%%%%%%%%%
\section{METHODS} \label{sec:methods}
The authors developed a two-level optimization framework \cite{ramasamy2022heterogenous} that uses Genetic Algorithm and Bayesian optimization and is described in Sec.~\ref{sec:two-level}. The main contribution of this paper is the A-teams architecture which uses the two-level optimization framework as its basis and is described in Sec.~\ref{sec:Ateams}.
%In order to solve this problem, a novel framework called A-Teams or Asynchronous Teams is proposed that gives high quality solutions with significantly reduced computing time. This framework keeps the underlying two-level optimization as the basis, but performs certain interactions between different optimization algorithms used. 

\subsection{Conventional Two-level optimization} \label{sec:two-level}
The Fig.~\ref{fig:algorithm} shows the conventional two-level optimization \cite{ramasamy2022heterogenous}. The  outer level block shown in blue are the heuristics to choose a UGV route. The UGV route heuristics has a few free parameters. Once these parameters are set, the inner-level block shown in orange performs the UAV route optimization using Google’s OR-Tools\texttrademark \cite{ORtools}. The UAV route optimization is heavily dependent on the free parameters of the UGV route. These parameters are optimized using a genetic algorithm. The complete block (inner/outer loop) runs several times till the genetic algorithm can no longer improve the solution or when the maximum iteration limit is reached.

%Thus far, the UAV-UGV route selection is open-loop since the UGV route has not been optimized. Thus, we run an optimization on the outer loop that optimizes the free parameters in the UGV heuristics minimizing an appropriate cost, thus closing the loop. This outer-inner loop optimization proceeds till the maximum iteration limit is reached or the solution has converged. That is, there is no change in the objective value.}
%
%\textcolor{blue}{Our heuristics for UGV route are based on maximum fuel range of the UAV described earlier. Figure ~\ref{fig:scenario_description} gives a graphical representation about how two-level optimization is implemented. The UGV starts from any one of those depots and travels along the mission points as shown by the arrow. Next, the UGV is allowed to stop anywhere in the ellipse with dashed red lines for a prescribed time. The rationale is that in choosing a stop and wait time is to give the UAV enough time to land and recharge on the UGV. Next, the UGV moves to another stop anywhere inside the blue ellipse with blue dash-dot lines. We have shown two random stop locations in each ellipse with a blue hollow circle. For the chosen stop locations and wait times for each stop, the UAV routing problem is formulated and solved.} 
%
%In order to prove the efficiency of this framework, the results obtained via this framework are compared with results of the conventional two-level optimization algorithm implemented in \cite{ramasamy2022heterogenous}. 

\subsection{Description of the proposed architecture - A-Teams} \label{sec:Ateams}
 A-Teams is an architecture that uses a team of autonomous agents to perform optimization on a given problem. The agents have a common set of potential solutions. Each agent works asynchronously on the potential solutions to find better solutions which are then updated as the new potential solutions. There are three main agents that constitute this architecture and are described next. 
 %Those are called {\bf Constructors}, {\bf Improvers} and {\bf Destroyers}.
 
{\bf 1. Constructor Agent} is used to develop an initial pool of solutions using the user inputs. 

{\bf 2. Improver Agent} is used to improve on the pool of solutions  using different optimization methods. It is important to choose complementary optimization methods (e.g., global and local optimizers) to help improve the quality of the solutions.  

{\bf 3. Destroyer Agent} is used to discard non-optimal and bad solutions. It does this by ranking the solutions based on the cost and constraints satisfaction. 
 
%These agents described above decides when to work and what solutions to work on by looking at the population of solutions and choosing a solution or a subset of solutions that contribute to the improvement of the optimization process.
% 
{\bf Populations} are shared repositories for storing solutions computed and evaluated by different agents. These are accessible by all agents.

The architecture is modular and distributed which enables each optimizer to work independently. However, the architecture also has mechanisms to combine solutions generated by individual optimization to realize further improvements of the solutions. This makes the framework very powerful producing optimal solutions in a computationally efficient manner. 

%Thus, each optimizer can work independently to 

%As it is known that the VRP is a NP-Hard combinatorial optimization problem \cite{braekers2016vehicle}, it is complex to obtain an optimal solution within a reasonable amount of time. This is due to the presence of several constraints in the problem. Standard heuristics in the literature can circumvent this by giving a near-optimal solution, but they are problem specific and might not provide the best solution for a certain set of problems. This is where this architecture comes in hand to have different agents in modules that uses different heuristic algorithms to work autonomously and co-operate together by sharing the population to provide the best solution efficiently. This way, the combined effort would produce solutions that are better than if produced by each of such algorithms.
% 
%The basic overview of the architecture is described above. The following subsection explains about how this architecture is implemented towards solving a co-operative routing problem.

%We now discuss how A-teams is used for producing solution for the co-operative routing problem. The proposed architecture overcomes the intricacies of planning an appropriate and optimal path for the UGV. \textcolor{blue}{One thing to note in this architecture is that, the implementation of the two-level optimization explained before is kept as the basis for solving this cooperative routing problem upon which different agents and algorithms work autonomously to provide a quick and quality solutions.} 

%\textcolor{red}{start here}
Figure ~\ref{fig:Architecture description} shows the implementation of A-Teams to solve this problem. Note that the A-teams operates over the two-level optimization block shown in Fig.~\ref{fig:algorithm}. 
The Constructor agent utilizes a `randomized' initial UGV parameter set to construct an initial population of solutions. The algorithm is used until the feasibility of a solution in the population is achieved. Every time the constructor agent pulls solution from that initial population, it fills up a current population, that was initially empty (represented in orange box in Figure ~\ref{fig:Architecture description}), until feasibility. These solutions are usually sub-optimal, but are then passed to the Improver agent. 
The Improver agents improve the solutions by using two algorithms: (1) the Nelder-Mead, a gradient free direct search method, that is good for  local optimization and (2) the Genetic Algorithm that is inspired from natural selection and is good global optimization method. These two algorithms are complementary in nature; Genetic algorithm searches big regions of the parameters set (exploration) while Nelder-Mead improves on the solution in the vicinity of the current solution (exploitation).

We now describe Algorithm \ref{algorithm:ATeams} used in A-teams.
On lines $1$ and $2$ the constructor agent role is to generate feasible solutions for the improver agent. A random initial parameter set for UGV is generated and checked if it leads to a feasible UAV solution. After a sufficient number of good feasible solutions are generated, the algorithm proceeds to the main while loop that uses the constructor and destroyer agent.  When the constructor agent produces a feasible solution, the current population which has been updated so far from the initial population is sorted, and the role is handed over to the Improver agents. The solutions are sorted based on the cost and the improver agent uses the global optimizer (GA) shown on line $5$ or local optimizer (Nelder-mead) shown on line $6$. These improver agents work in parallel. Thus, both, the exploration and the exploitation happens simultaneously and independently. Next, on line $7$, the destroyer agent looks at all the solutions and discards the infeasible solutions and those that are already existing in the pool of solutions. Finally, on line $8$, all the good solutions are pooled together and sorted in order to get ready for the next iteration. This process repeats until convergence is achieved, i.e., when there is no improvement in the population.

%Thus, the optimization performed by the agents in the architecture attribute to the optimal free parameters in the UGV heuristics which is utilized to perform inner-level optimization of UAV route.

\begin{algorithm} [t]
    \caption{A-Teams architecture} 
  \begin{algorithmic}[1]
    \INPUT Population size, $n$; Initial population
    \OUTPUT Global best solution
    \STATE \textbf {Constructor agent:} Generate random initial population with population size $n$;
    
    \STATE \textbf {Constructor agent:} Perform the UAV optimization for corresponding UGV parameter set until feasibility;
    %\STATE Compute the fitness value for each chromosomes
    \WHILE {Convergence is not achieved}
    	\STATE The following two improver agents work in parallel 
	\STATE \parbox[t]{200pt}{\textbf {Improver agent 1:} Perform Nelder-Mead optimization for local improvement on the current best solution; \strut}
	\STATE \parbox[t]{200pt}{\textbf {Improver agent 2:} Perform Genetic Algorithm optimization for global improvement on the current population; \strut}
	\STATE \parbox[t]{200pt}{\textbf {Destroyer agents 1 and 2:} Destroy the infeasible or already existing solutions on the fly; \strut}
	\STATE \parbox[t]{200pt}{Replace initial or old population with newly generated population; \strut}
	\STATE Compute the fitness value for each population member and sort them in ascending order;
	%\STATE Increment the current generation $g$ by 1;
	%\ENDIF
    \ENDWHILE
  \end{algorithmic}
  \label{algorithm:ATeams}
\end{algorithm}

\subsection{Heuristics for UGV (Outer-level)} 

Our heuristics for UGV route are based on maximum fuel range of the UAV described earlier (range is shown as a blue circle in Fig.~\ref{fig:scenario}). Figure~\ref{fig:scenario_description} shows the heuristics for the UGV route. The UGV starts at the depot and travels along the target points. Next, the UGV is allowed to stop anywhere in the ellipse with dashed red lines for a prescribed time. The rationale is that in choosing a stop and wait time is to give the  UAV enough time to land and recharge on the UGV. Next, the UGV moves to to the bottom right side and can take another stop anywhere inside the blue ellipse with blue dash-dot lines. We have shown two random stop locations in each ellipse with a blue hollow circle. There are 7 parameters for the UGV heuristics; the starting location of the UGV/UAV, the x- and y-coordinate of each of the two stop locations, and the wait times at the stops.

%%%%%%%%%%%%%%%%%%%%%%%%%%%%%%%
\subsection{Optimizing UAV route (Inner-level)}
We formulate a Vehicle Routing Problem (VRP) with capacity constraints to account for fuel limits, time windows to allow for rendezvous, and dropped visits to allow the UAV to visit some of the many vertices on the UGV path. We constrain the UAV to a fixed speed, pre-specify the battery capacity and service time as the UAV lands and waits on the UGV. Constrained Programming approach is being used to solve this VRP using OR-Tools solver.

The mathematical details are note included here because of space constraint, but can be found in \cite{ramasamy2022heterogenous}.

\section{RESULTS} \label{sec:results}
We used Python 3 for all the computations: a custom-written genetic algorithm and Nelder Mead from Scipy package for UGV parameter optimization, and OR-tools for UAV optimization. All computations were done on a 3.7 GHz Intel Core i9 processor with 32 GB RAM on a 64-bit operating system.

In order to prove the computational efficiency, we present the results on three different scenarios shown in Fig.~\ref{fig:scenario}. The scenarios under consideration have three branches that intersect at a single point. Each scenario has three depots. At each of the these depots, the UAV or the UGV may be recharged. The UAV may also recharged when it lands on the UGV. The UGV/UAV start their route execution from one of the three depots. This location is one of the free UGV parameter. All these scenarios consider the optimization problem for 1 UGV and 1 UAV. The UAV and UGV velocities when moving are fixed at 10 m/s and 4 m/s respectively. The UAV and UGV fuel capacity are 287.7 kJ and 25.01 MJ respectively. 

%%%%%%%%%%%%%%%%%%%%%%%%%%%%%%%
\begin{table}[tb]
\centering
\begin{tabular}{| p{1.5 cm} | p{1.5 cm}| p{1.5 cm}| p{1.5 cm}|} 
 \hline
 Parameter & \multicolumn{3}{c|}{Range} \\ [1ex] 
 \hline\hline
 &  Scenario 1 & Scenario 2 & Scenario 3 \\[1ex]
 \hline\hline
 UGV stop 1 (km,km) &   (6.02,16.82) to (4.99, 11.65)     & (11.36,4.86) to (14.34, 7.31) & (8.95,9.48) to (9.29, 9.38) \\ [1ex] 
 UGV stop 2  (km,km) &   (14.70, 4.02) to (16.96,1.45) & (7.72,6.13) to (9.46, 13.02) & (8.61,9.82) to (8.61, 10.07) \\ [1ex] 
 UGV wait 1,2 (min) & 2 to 50 & 2 to 50 & 2 to 50 \\ [1ex] 
Starting point                 & 1,2, or 3 & 1,2, or 3 & 1,2, or 3 \\[1ex]
\hline
\end{tabular}
\caption{UGV parameters and their ranges (outer loop)}
\label{table:UGV_parameters}
\end{table}

%%%%%%%%%%%%%%%%%%%
\begin{figure*} [h]
\centering
\includegraphics[scale=0.43]{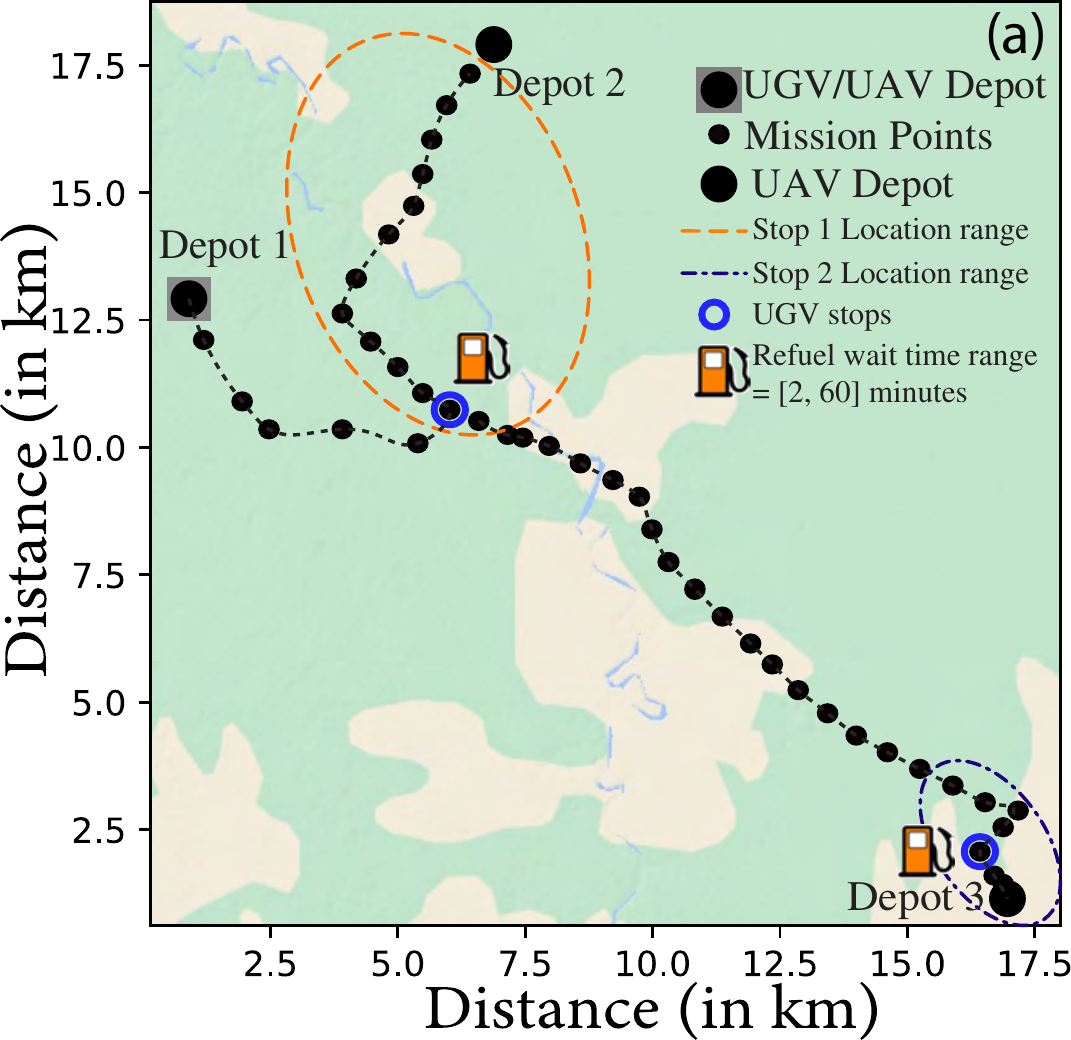}
\includegraphics[scale=0.43]{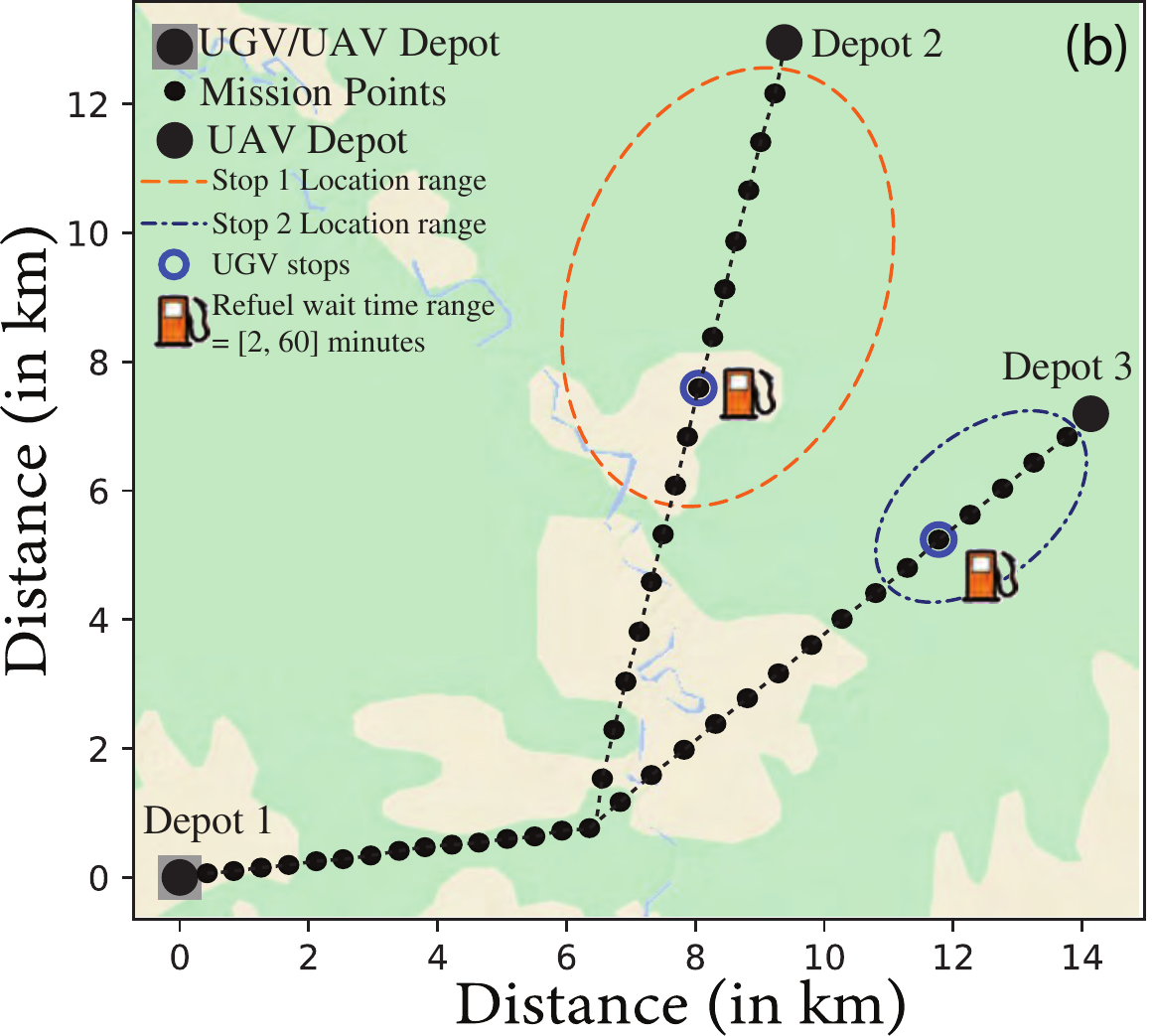}
\includegraphics[scale=0.43]{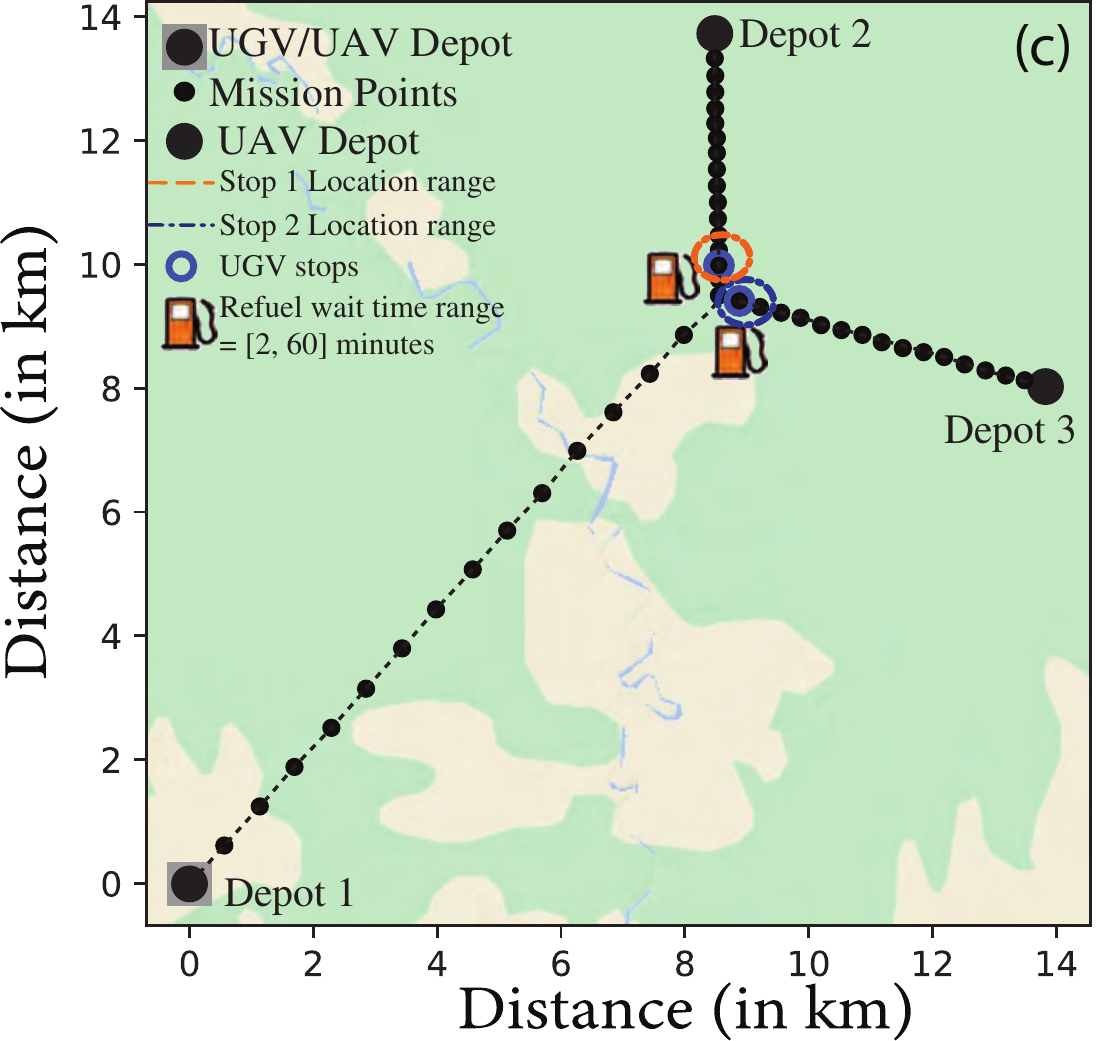}
\vspace{-0.25cm}
\caption{Description of different scenarios with parameters to be optimized. a) Scenario 1 b) Scenario 2 c) Scenario 3.}
\label{fig:scenario_description}
\end{figure*}
%%%%%%%%%%%%%%%%%%%

%%%%%%%%%%%%%%%%%%%
%\begin{figure*} [htb]
%\centering
%\includegraphics[scale=0.5]{figures/Scenario 1 Desc.pdf}
%\vspace{-0.25cm}
%\caption{Scenario 2}
%\label{fig:scenario2}
%\end{figure*}
%%%%%%%%%%%%%%%%%%%

%%%%%%%%%%%%%%%%%%%
%\begin{figure*} [htb]
%\centering
%\includegraphics[scale=0.5]{figures/Scenario 2 Desc.pdf}
%\vspace{-0.25cm}
%\caption{Scenario 3}
%\label{fig:scenario3}
%\end{figure*}
%%%%%%%%%%%%%%%%%%%

%%%%%%%%%%%%%%%%%%%%%%%%%%%%%%%
\begin{table}[tb]
\centering
\begin{tabular}{| p{1.2 cm} | p{1 cm}| p{1 cm}| p{1 cm}| p{1 cm}|} 
 \hline
 Scenario type & \multicolumn{2}{p{1.75 cm}|}{Computational time (in min.)} & \multicolumn{2}{p{1.75 cm}|}{Objective (in min.)}\\ [1ex] 
 \hline\hline
 & A-Teams & Two-level optimization & A-Teams & Two-level optimization \\[1ex]
 \hline\hline
 Scenario 1 & $37 \pm 1$ & $47 \pm 2$ & 163 & 166 \\ [1ex] 
 Scenario 2 & $28 \pm 9$ & $82 \pm 10$ & 12 & 9 \\ [1ex] 
 Scenario 3 & $13 \pm 3$ & $44 \pm 2$ & 18 & 13 \\ [1ex] 
\hline
\end{tabular}
\caption{Comparison of total cost between A-Teams and conventional two-level optimization for different scenarios}
\label{table:comparizon}
\end{table}

%%%%%%%%%%%%%%%%%%%%%%%%%%%%%%%
%\begin{table}[tb]
%\centering
%\begin{tabular}{| p{2.1 cm} | p{2 cm}| p{2 cm}|} 
 %\hline
% Scenario type & \multicolumn{2}{c|}{Total mission time (in min.)} \\ [2ex] 
 %\hline\hline
 %& A-Teams & Two-level optimization \\[2ex]
 %\hline\hline
 %Scenario 1 & 228 & 228 \\ [2ex] 
 %Scenario 2 & 280 & 238 \\ [2ex] 
 %Scenario 3 & 136 & 136 \\ [2ex] 
%\hline
%\end{tabular}
%\caption{Comparison of total cost between A-Teams and conventional two-level optimization for different scenarios. Each scenario has the results for a specific initialization.}
%\label{table:costs}
%\end{table}

%%%%%%%%%%%%%%%%%%%%%%%%%%%%%%%%%%
\begin{table*}[tb]
\centering
\begin{tabular}{|p{3cm} | p{1.2cm} | p{1.2cm}| p{1.2cm}| p{1.2cm}| p{1.2cm}| p{1.2cm}|} 
 \hline
 Parameter & \multicolumn{6}{c|}{Optimal parameter values} \\ %[1ex] 
 \hline\hline
 &  \multicolumn{2}{c|}{Scenario 1} & \multicolumn{2}{c|}{Scenario 2} & \multicolumn{2}{c|}{Scenario 3} \\ %[1ex]
 \hline\hline
 & A-Teams & Two-level  & A-Teams & Two-level  & A-Teams & Two-level  \\[1ex]
 \hline
 UGV stop 1 location (km,km) & (4.99,11.65) & (4.99,11.65) & (7.92,6.90) & (11.36,4.86) & (8.61,10.08) & (8.95,9.48) \\ [1ex] 
 UGV stop 2 location (km,km) & (16.96,1.45) & (16.96,1.45) & (12.36,5.68) & (8.30,8.43) &(9.29,9.38) & (8.61,10.08) \\ [1ex] 
 UGV stop 1 wait time (min) & 20 & 20 & 50 & 21 & 20 & 22 \\ [1ex] 
 UGV stop 2 wait time (min) & 20 & 21 & 20 & 21 & 20 & 20 \\ [1ex] 
 Route starting and ending point                 & Depot 1& Depot 1 & Depot 3 & Depot 3 & Depot 3 & Depot 2 \\[1ex]
\hline
 Metrics &  \multicolumn{2}{c|}{Scenario 1} &  \multicolumn{2}{c|}{Scenario 2} &  \multicolumn{2}{c|}{Scenario 3} \\
 \hline
  &A-Teams  & Two-level &A-Teams & Two-level &A-Teams & Two-level \\ %[2ex] 
       %    &   & Optimization &   & Optimization &   & Optimization \\ [1ex] 
\hline\hline
Objective function (min) & 163 & 166 & 12 & 9 & 18 & 13 \\[1ex]
 Total time (min) & 228  & 231 & 216 & 201 & 145 & 131  \\ [1ex] 
 %%%%%%%%%%%%%%%%%
  \hline\hline
 UGV results & & & & & & \\ [1ex]
 \hline
Travel time (minutes) & 228  & 231 & 216 & 201 & 145 & 131 \\ [1ex] 
 Energy consumed (MJ) & 23.16 & 23.19 & 19.50 & 20.89 & 8.16 & 6.31 \\ [1ex] 
 \# Targets visited & 34 & 34 & 23 & 25 & 17 & 17 \\ [1ex] 
 \hline\hline
 UAV results &  & & & & & \\ [1ex]
 \hline
 Travel time (minutes) & 65  & 65 & 204 & 192 &127 & 118 \\ [1ex] 
  Energy consumed (kJ) & 460.65 & 460.65 & 1082.92 & 1301.78 & 841.72 & 874.73 \\ [1ex] 
 Recharging stops on UGV & 1 & 1 & 2 & 3 & 2 & 2 \\ [1ex] 
  Recharging stops on Depot & 0 & 0 & 2 & 2 & 1 & 1 \\ [1ex] 
\# Targets visited  & 10 & 10 & 23 & 21 & 29 & 29 \\ [1ex] 
%%%%%%%%%%
\hline
\end{tabular}
\caption{Comparison between A-Teams and conventional two-level optimization on metrics from the optimal solution for different scenarios. These results are shown for a specific initialization of population by Constructor agent.}
\label{table:UAV and UGV optimal solutions}
\end{table*}
%%%%%%%%%%%%%%%%%%%%%%%%%%%%%%%%%%

%\begin{table*}[t]
%\centering
%\begin{tabular}{| p{2.5 cm} | p{1.5 cm}| p{1.5 cm}| p{1.5 cm}| p{1.5 cm}| p{1.5 cm}| p{1.5 cm}|} 
 %\hline
% Parameter & \multicolumn{6}{c|}{Optimal parameter values} \\ [1ex] 
 %\hline\hline
 %&  \multicolumn{2}{c|}{Scenario 1} & \multicolumn{2}{c|}{Scenario 2} & \multicolumn{2}{c|}{Scenario 3} \\[1ex]
 %\hline\hline
 %& A-Teams & Two-level optimization & A-Teams & Two-level optimization & A-Teams & Two-level optimization \\[1ex]
 %\hline
 %UGV stop 1 location (km,km) & (4.99,11.65) & (4.99,11.65) & (8.30,8.43) & (11.86,5.26) & (8.61,9.82) & (9.29,9.38) \\ [1ex] 
 %UGV stop 2 location (km,km) & (16.96,1.45) & (16.96,1.45) & (12.84,6.08) & (8.11,7.66) &(8.95,9.48) & (8.61,9.82) \\ [1ex] 
 %UGV stop 1 wait time (min) & 20 & 20 & 40 & 23 & 20 & 20 \\ [1ex] 
 %UGV stop 2 wait time (min) & 20 & 21 & 49 & 21 & 20 & 20 \\ [1ex] 
 %Route starting point                 & Depot 1& Depot 1 & Depot 3 & Depot 3 & Depot 3 & Depot 3 \\[1ex]
%\hline
%\end{tabular}
%\caption{Optimal parameters after optimization via respective algorithms}
%\label{table:UGV optimal solutions}
%\end{table*}

Figure~\ref{fig:scenario_description} shows the UGV parameters for the three scenarios. The black dot on the gray rectangle represents the depot where both UGV and UAV can recharge. The large black circles represents the locations where only the UAV can recharge. Either of those depots represent the potential starting location for the UAV and UGV and is an optimization parameter. The small black circles represent the target points that need to be visited either by the UGV or the UAV. The stopping locations for the UGV can be either in the red ellipse or the blue ellipse. In each of this ellipse, the x- and y-coordinate is a parameter (2 parameters per ellipse). For each stop location, the wait time is also a free parameter (1 parameter per ellipse). The UAV/UGV may start at Depot 1, 2, or 3 (1 parameter). Table ~\ref{table:UGV_parameters} shows the UGV parameter range for the outer level. The objective function is to minimize the time gap between completion of UAV's and UGV's routes after visiting all their target points in the respective scenario. This kind of objective function helps to minimize the waiting time between the heterogeneous system after the route execution cycle.

%The parallel computing nature of the A-Teams architecture is to provide a computationally efficient solution that is near-optimal. In order to have a fair comparison between the proposed algorithm and the conventional two-level optimization algorithm exploited from the previous work \cite{ramasamy2022heterogenous}, parallel computing is performed on the latter as well. 

%In order to show the robustness of the architecture, different random initialization are thrown into the initial solution of these two algorithms and the results show the mean and standard deviation values of the computational time. The initial population size is considered to be $n=30$ for both the algorithms. In case of two-level optimization algorithm the maximum number of generations in the genetic algorithm, which is used to optimize UGV route, is considered to be $g_{max}=11$. This high number of generations in the genetic algorithm allows to achieve convergence towards the optimal solution.

In order to compare the two methods, an initial population size of $30$ was chosen and both algorithms were run $3$ times. Table~\ref{table:comparizon} compares the cost and the computational time. It can be seen that the computational time is reduced by a factor of $2$ to $3$ by the A-teams architecture in comparison to two-level optimization, but the cost is within $25 \%$.

Table~\ref{table:UAV and UGV optimal solutions} compares the A-teams solution with two-level optimization for the same initial population for the three different scenarios. From the value of the objective in the table it can be seen that the results are mixed: A-teams is better than two-level optimization for scenario 1 but not for scenario 2 and 3. However, the difference between the two is not substantially large. The other metrics such as the UAV/UGV travel time, energy consumed, recharging stops, targets visited are also shown. There are minor differences between the two. 

\begin{figure*} [h]
\centering
\includegraphics[scale=0.5]{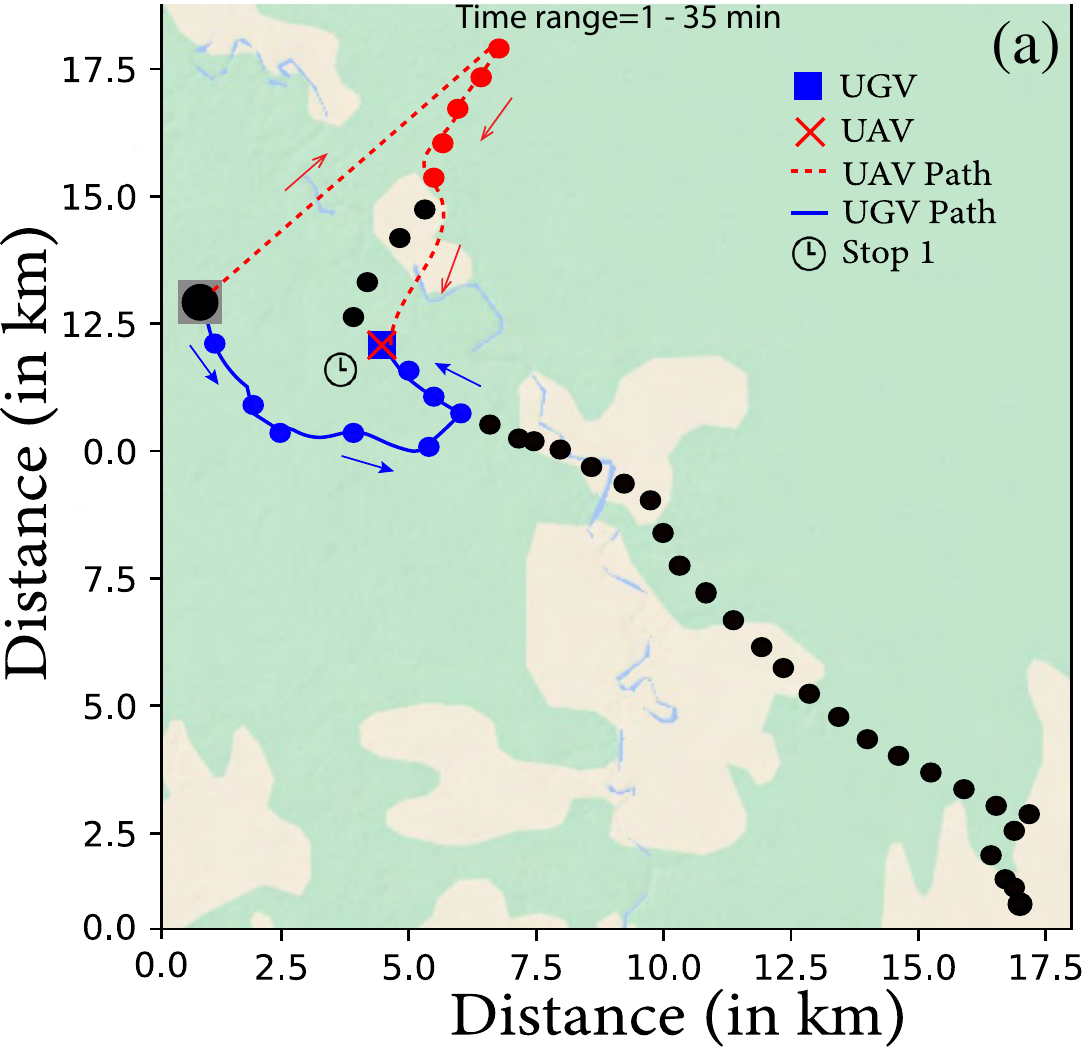}
\includegraphics[scale=0.5]{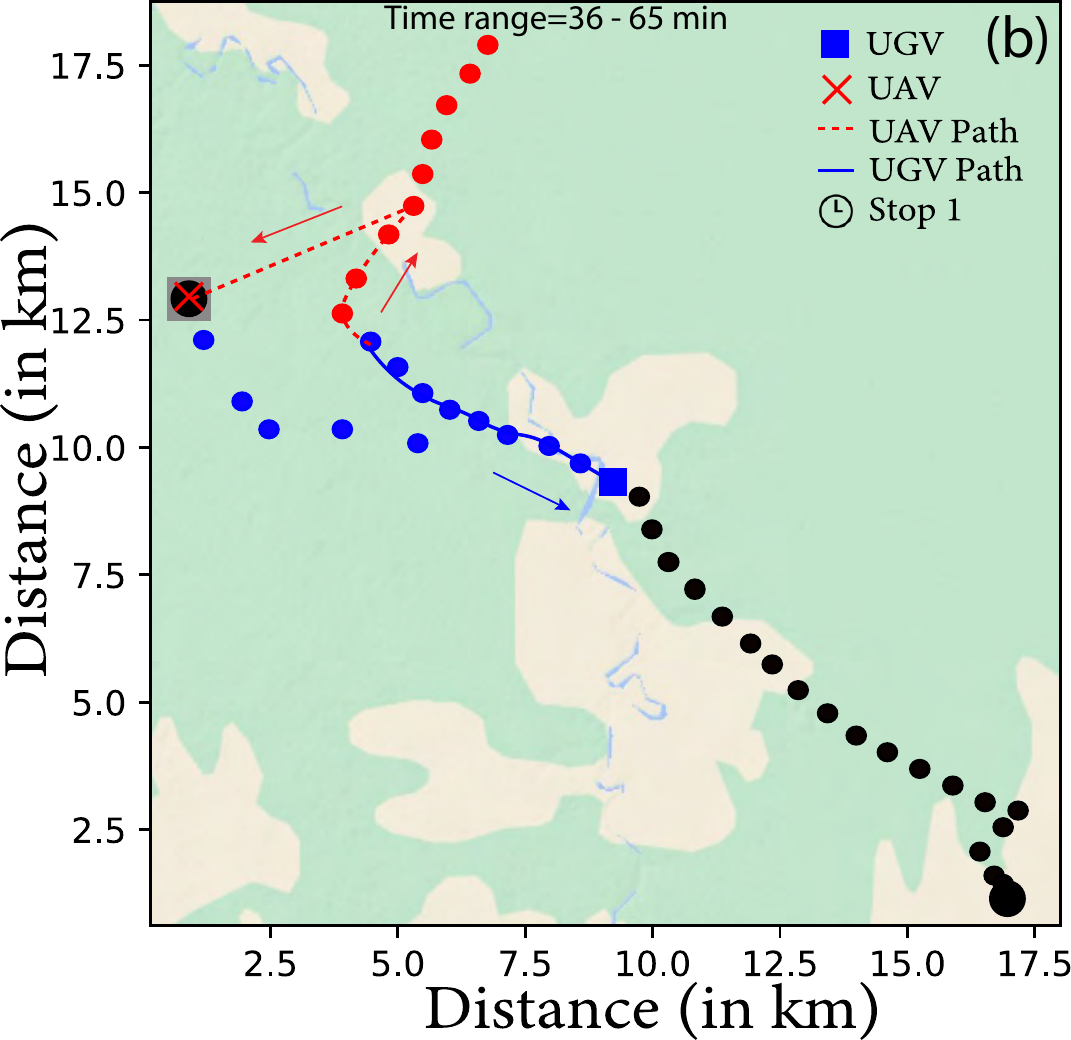}
\includegraphics[scale=0.5]{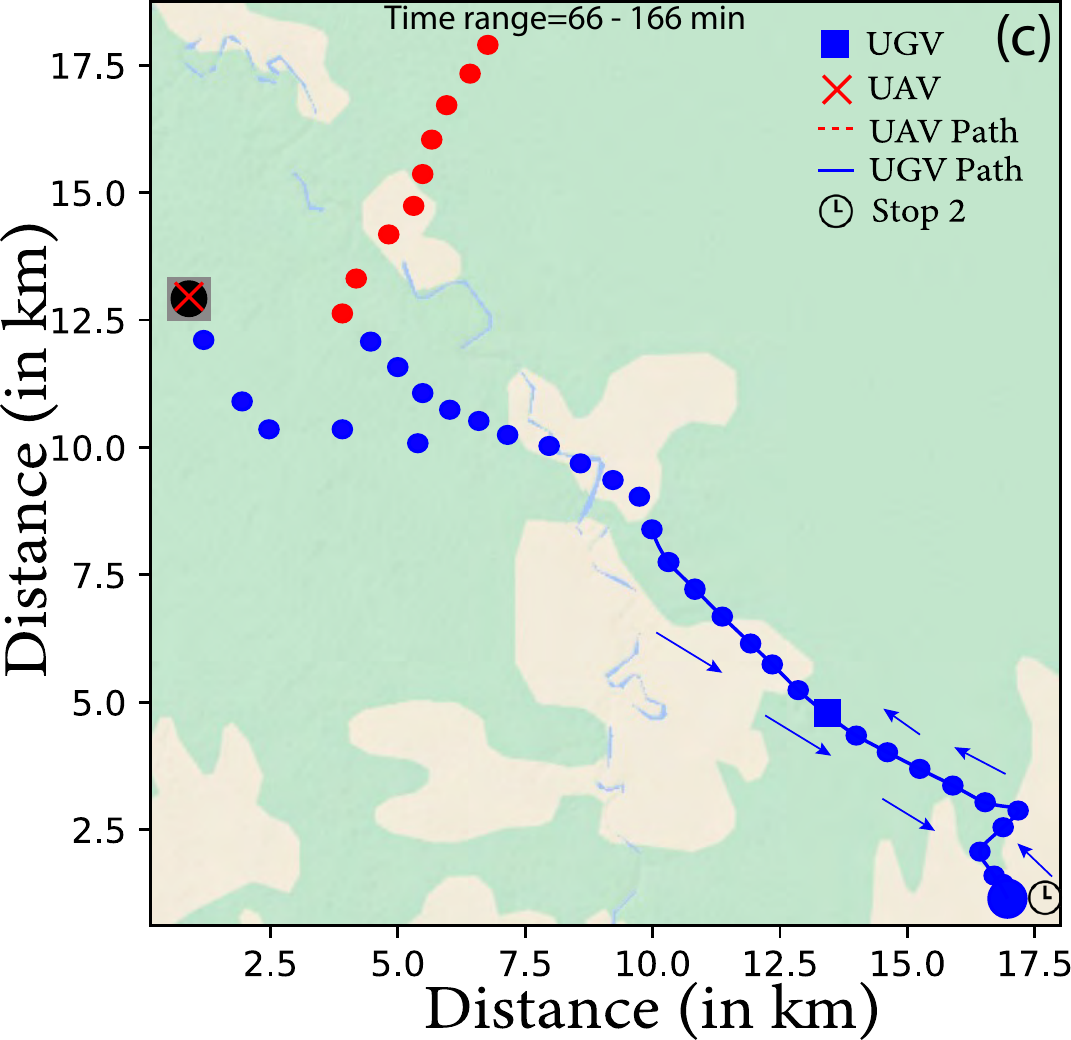}
\vspace{-0.25cm}
\caption{Solution produced by conventional two-level optimization and A-teams on Scenario 1 are indistinguishable and are shown here. The different plot shows the UAV and UGV route at various time-steps.}
\label{fig:GA_LS_timestep_route}
\end{figure*}

\begin{figure*} [h]
\centering
\includegraphics[scale=0.5]{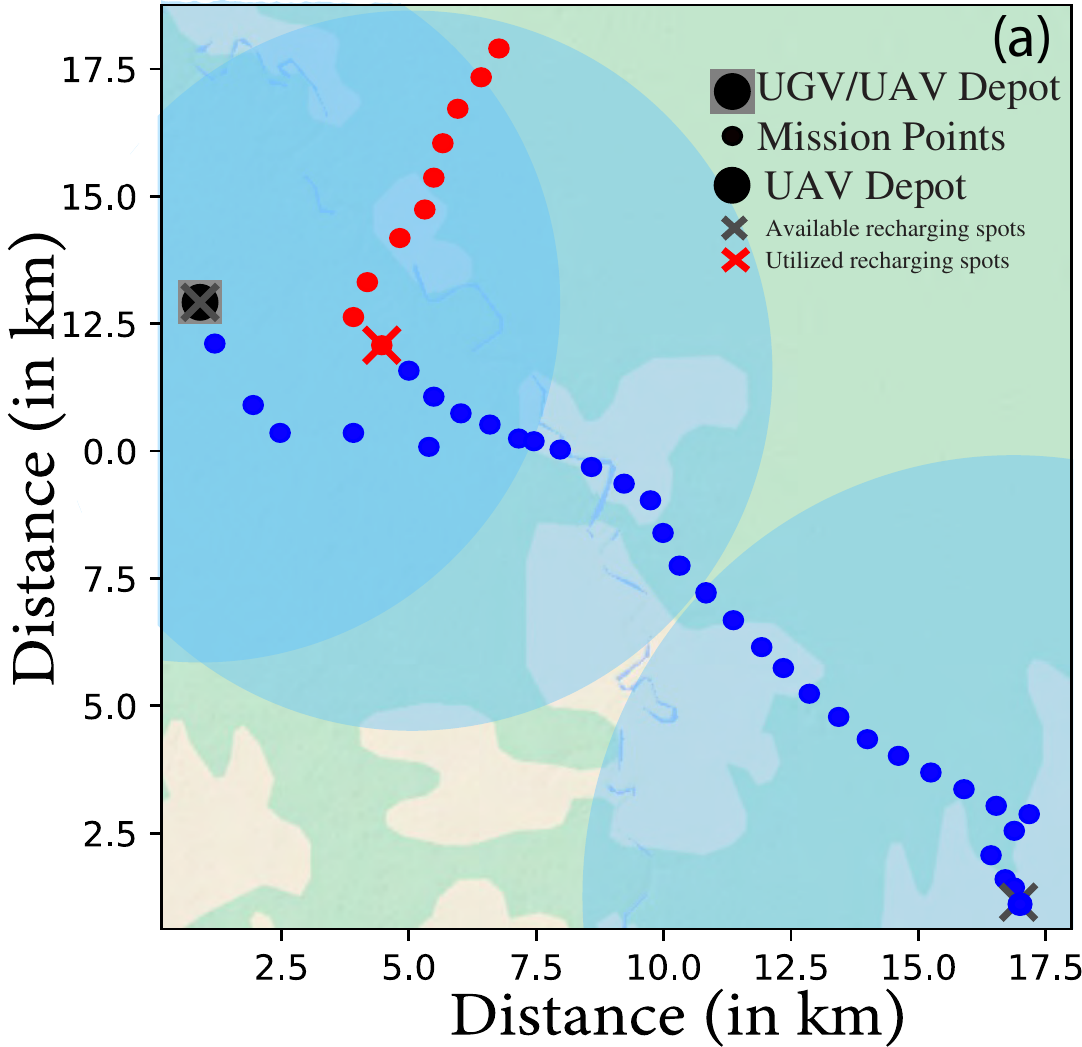}
\includegraphics[scale=0.5]{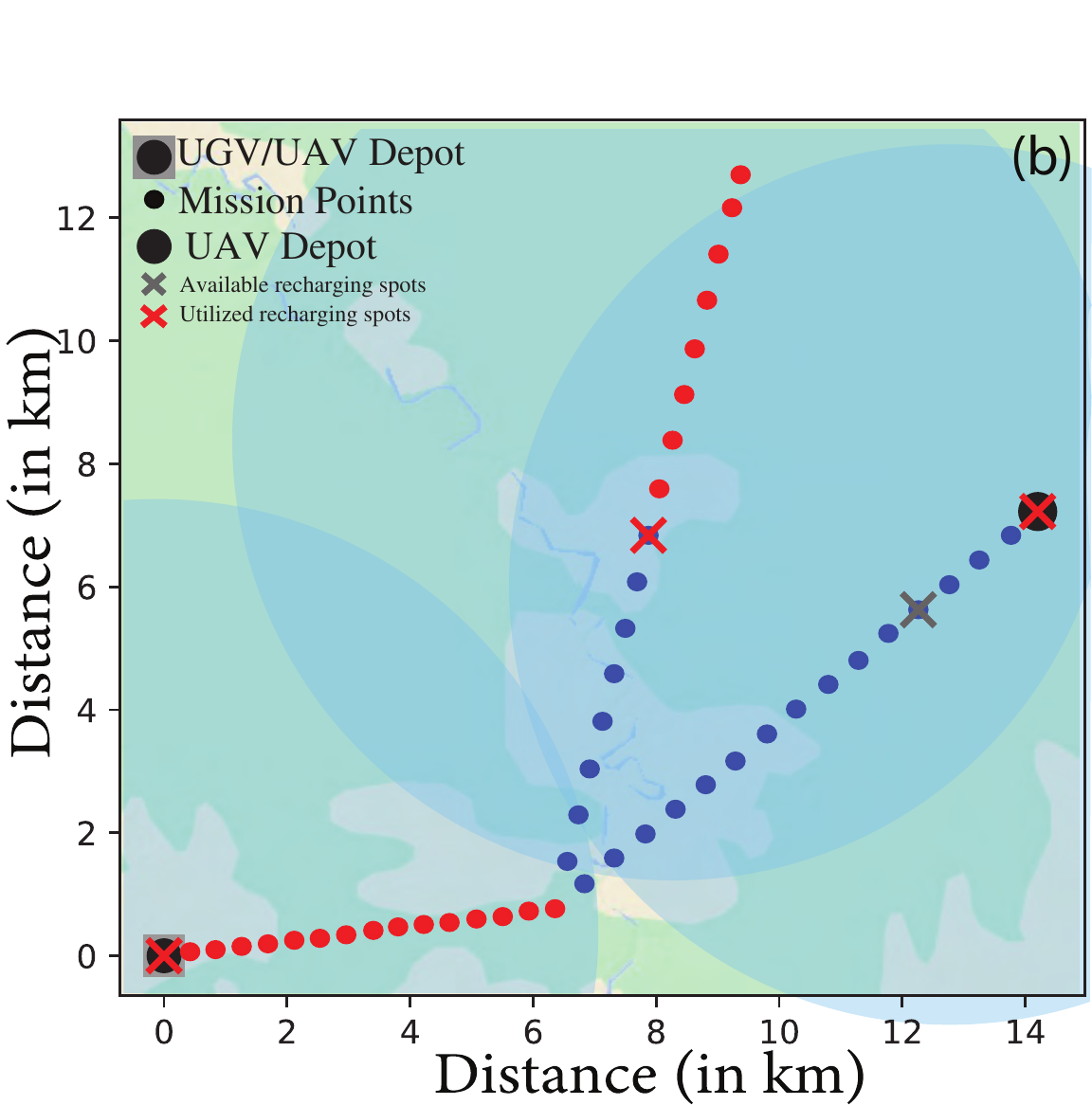}
\includegraphics[scale=0.5]{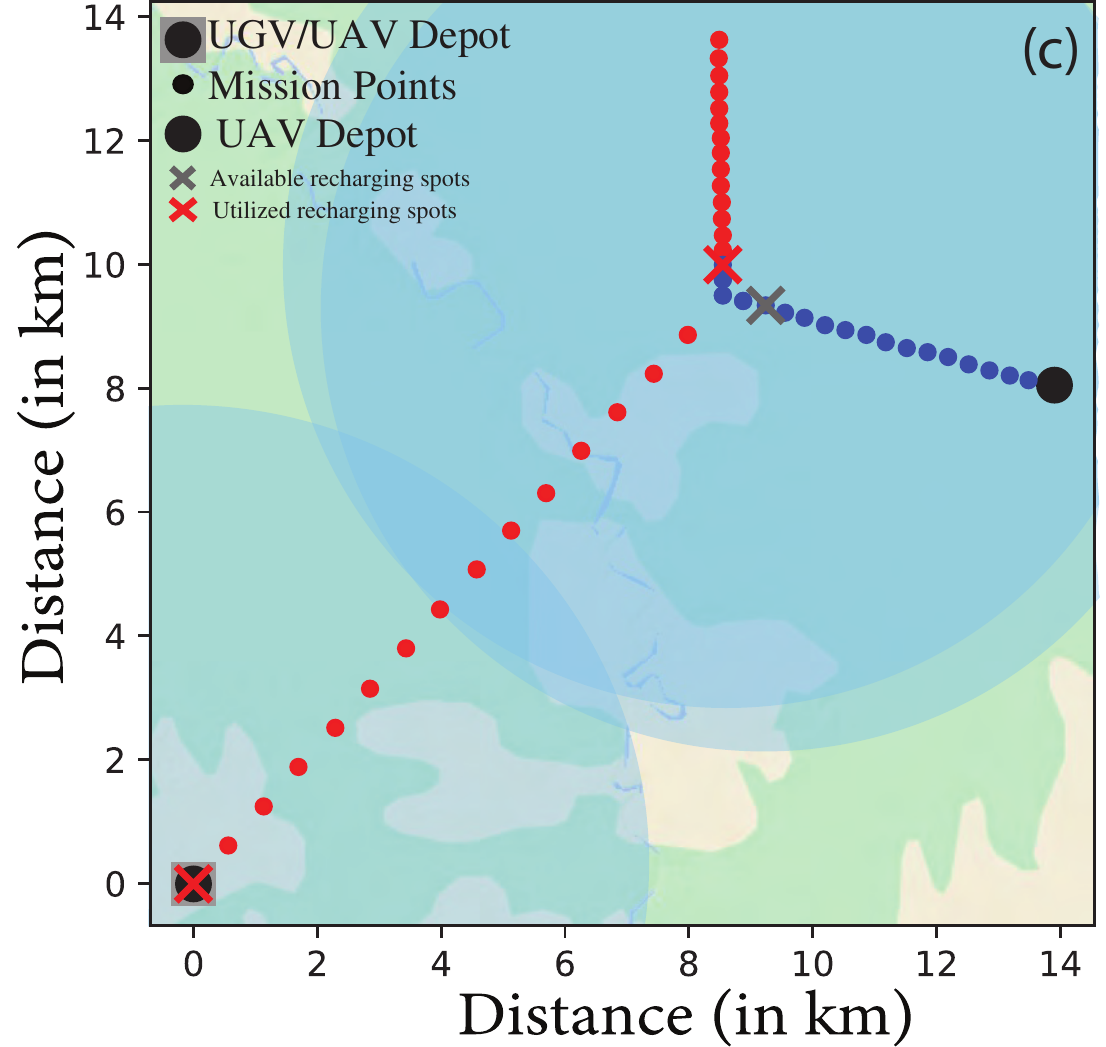}
\vspace{-0.25cm}
\caption{Optimal parameter results of respective scenarios obtained using the A-Teams architecture. (a) Scenario 1 (b) Scenario 2 (c) Scenario 3}
\label{fig:scenario_results}
\end{figure*}

%Figure~\ref{fig:A-Teams_timestep_route} and 
Figure~\ref{fig:GA_LS_timestep_route} shows the final solution for scenario 1 at three different time ranges (in min): $1-35$, $36-65$, and $66-166$. Since both two-level and A-teams produce very similar solution, only one solution, the two-level optimization, is shown here. The total routing time for A-teams is less than that of the two-level optimization by about $3$ min. The UAV/UGV start at Depot 1 then they move together to the first stop location. Here the UAV flies to cover the targets on the top portion returning to Depot 1 to recharge. Then the UGV travels to all the target points and returns back to the Depot 1. Fig.~\ref{fig:scenario_results} shows the coverage of target points and the recharging stops used by the UAV-UGV for the A-teams for scenario 1. The target points in red are those that are covered by the UAV while those in blue are the ones covered by the UGV. The red cross shows the stopping locations for the UAV on the UGV for recharging. The overlaid light blue circles indicate the range of the UAV. It can be seen that the recharging stops are chosen strategically to enable maximum fuel coverage for the UAV on a single charge.  
%%%%%%%%%%%%%%%%%%%%
%\begin{figure*} [h]
%\centering
%\includegraphics[scale=0.5]{figures/A_teams_res_1.pdf}
%\includegraphics[scale=0.5]{figures/A_teams_res_2.pdf}
%\includegraphics[scale=0.5]{figures/A_teams_res_3.pdf}
%\vspace{-0.25cm}
%\caption{Solution produced by A-Teams on Scenario 1. The different plot shows the UAV and UGV route at various time-steps}
%\label{fig:A-Teams_timestep_route}
%\end{figure*}
%%%%%%%%%%%%%%%%%%%%

%%%%%%%%%%%%%%%%%%%
%%%%%%%%%%%%%%%%%%%

%%%%%%%%%%%%%%%%%%%
\section{DISCUSSION} \label{sec:discussion}
This paper presented the A-teams framework for optimizing the routes of a UAV-UGV pair subject to fuel and speed constraints. The A-teams framework uses asynchronous agents to create an initial pool of solutions, improve the pool, and then destroy the infeasible solutions. These agent exploit parallel architecture to produce fast solutions. When compared with conventional optimization method, A-teams produces the solution $2-3$ times faster while achieving similar quality of solutions. 

One advantage of A-teams is the use of asynchronous agents to improve the solutions. These agents work in parallel and hence they can be deployed independent of each other. When the algorithm is deployed either on multiple core or parallel computing machines, they are able to speed up computations. 

Another advantage of A-teams is that the architecture seamlessly exploits the advantages of multiple algorithms to improve the solution. In our case, the genetic algorithm is used to explore the search space while the Nelder-Mead is used to locally improve the solution. Thus, we have combined a global search with local search to improve the solution quality. However, genetic algorithm is not sample efficient. One could use a sample efficient method like Bayesian Optimization if sample efficiency is important \cite{ramasamy2022heterogenous}. 

The proposed works has some disadvantages. The UGV heuristics, the stop location and wait times, were manually determined by hand tuning. This may be overcome by using minimum set cover algorithm \cite{maini2019cooperative}. The quality of the initial pool of solutions created by the constructor agent is critical to ensure that the improver agent is able to improve the solution. Thus, we had to play with a few random initial guesses till we got a feasible solution as a starting base. Our results indicate that the A-teams produce superior solutions for complex scenarios (Scenario 1), but was unable to produce better solutions that our baseline method of using genetic algorithms in simple scenarios (Scenario 2 and 3). This might indicate that the more complex A-teams architecture might not be ideal for certain scenarios.

\section{CONCLUSIONS AND FUTURE WORK} \label{sec:conclude}
We conclude that Asynchronous multi-agent architecture (A-teams) is a competitive tool for solving cooperative heterogeneous Vehicle Routing Problem problem. A-teams is able to produce good quality solutions in a computationally less time. It is able to do so by using specialized agents: agents to create solutions, agents to improve solutions globally and locally, and agents to destroy bad solutions. 

Our future work will explore methods to automate the choosing of UGV parameters, testing the scalability of the approach by adding more UGV parameters, more UAVS and UGVs, and testing other algorithms such as Bayesian or reinforcement learning to improve the quality of the solutions as well as the solution time.  
%Given the faster computation of the solutions, our future work would increase the number of UAVs considered and also explore the possibilities of applying Reinforcement Learning (RL) to automate the optimization process by training different scenarios based on just the scenario description inputs. 

\section*{Acknowledgment}
The authors would like to thank  Jean-Paul F. Reddinger, James M. Dotterweich, and Marshal A. Childers from DEVCOM Army Research Laboratory, Aberdeen Proving Grounds, Aberdeen, MD 21005 USA and James D. Humann is with DEVCOM Army Research Laboratory, Los Angeles, CA, 90094 USA for providing solutions that helped improve the formulation and solutions.

%   $^2$ Jean-Paul F. Reddinger, James M. Dotterweich, Marshal A. Childers are with DEVCOM Army Research Laboratory, Aberdeen Proving Grounds, Aberdeen, MD 21005 USA. {\tt\small jean-paul.f.reddinger.civ@army.mil}   
% \hspace{2mm}
% \hspace{2mm}  {\tt\small james.m.dotterweich.civ@army.mil} \hspace{2mm}  {\tt\small marshal.a.childers.civ@army.mil}. $^3$ James D. Humann is with DEVCOM Army Research Laboratory, Los Angeles, CA, 90094 USA. {\tt\small  james.d.humann.civ@army.mil}  

\bibliographystyle{plain}   % this means that the order of references
			    % is dtermined by the order in which the
			    % \cite and \nocite commands appear			    
\bibliography{ICUAS_2023_final.bib}  % list here all the bibliographies that
			     % you need. 
\end{document}